\documentclass[runningheads]{llncs}

\usepackage[T1]{fontenc}
\usepackage{amsmath}
\usepackage{tablefootnote}
\usepackage{threeparttable}
\usepackage{graphicx}
\usepackage{multirow}
\usepackage{array}
\usepackage{tikz}
\usepackage{xcolor}
\usepackage{tcolorbox}
\usepackage{amssymb}

\usepackage{longtable}
\usepackage{diagbox}
\usepackage{tikz-qtree}
\usepackage[edges]{forest}
\usetikzlibrary{trees} 
\usepackage{hyperref}

\begin{document}
\title{Privacy in Fine-tuning Large Language Models: Attacks, Defenses, and Future Directions}
\titlerunning{Privacy in Fine-tuning LLMs: Attacks, Defenses, and Future Directions}
%
\author{Hao Du\inst{1} \and Shang Liu\inst{2} \and Lele Zheng\inst{3} \and Yang Cao\inst{3} \and Atsuyoshi Nakamura\inst{1} \and Lei Chen\inst{4}}
\authorrunning{H. Du \and S. Liu et al.}
\institute{Hokkaido University \and China University of Mining and Technology \and Institute of Science Tokyo \and Hong Kong University of Science and Technology}
%
%
\maketitle              
\begin{abstract}

Fine-tuning has emerged as a critical process in leveraging Large Language Models (LLMs) for specific downstream tasks, enabling these models to achieve state-of-the-art performance across various domains. 
However, the fine-tuning process often involves sensitive datasets, introducing privacy risks that exploit the unique characteristics of this stage. 
In this paper, we provide a comprehensive survey of privacy challenges associated with fine-tuning LLMs, highlighting vulnerabilities to various privacy attacks, including membership inference, data extraction, and backdoor attacks. 
We further review defense mechanisms designed to mitigate privacy risks in the fine-tuning phase, such as differential privacy, federated learning, and knowledge unlearning, discussing their effectiveness and limitations in addressing privacy risks and maintaining model utility. 
By identifying key gaps in existing research, we highlight challenges and propose directions to advance the development of privacy-preserving methods for fine-tuning LLMs, promoting their responsible use in diverse applications.

\keywords{Privacy  \and Fine-tuning LLMs  \and Attacks \and Defenses.}
\end{abstract}
\section{Introduction}

In recent years, the rapid advancement of large language models (LLMs) such as GPT-4, LLaMA, and PaLM has revolutionized natural language processing (NLP), enabling applications across diverse domains such as content generation and code completion. These models are trained on vast amounts of data, often sourced from publicly available text on the Internet, to achieve remarkable performance in understanding and generating human-like text. However, the widespread deployment of LLMs also raises significant privacy concerns, especially as they become increasingly integrated into sensitive applications, such as healthcare, finance, and customer service.

Fine-tuning has emerged a highly popular approach for adapting pre-trained large language models (LLMs) to specific downstream tasks. By leveraging pre-trained knowledge and tailoring it to domain-specific needs, fine-tuning enables enhanced model performance on specialized applications and domains like autonomous systems, legal applications, and healthcare.
However, this process often involves sensitive information, such as sensor data, confidential legal documents, or personal medical records, raising significant privacy concerns.
In recent years, numerous novel fine-tuning techniques have been developed \cite{ben-zaken-etal-2022-bitfit,hu2022lora,lester-etal-2021-power,li-liang-2021-prefix,xu2023parameter}. Beyond the traditional full fine-tuning~\cite{devlin2018bert}, methods like Parameter-Efficient Fine-Tuning (PEFT)~\cite{han2024parameter} have gained traction, significantly reducing the resource overhead, such as time and computational cost. 
Conversely, the rapid development of new fine-tuning methods and their diverse applications have also introduced new attack surfaces. 

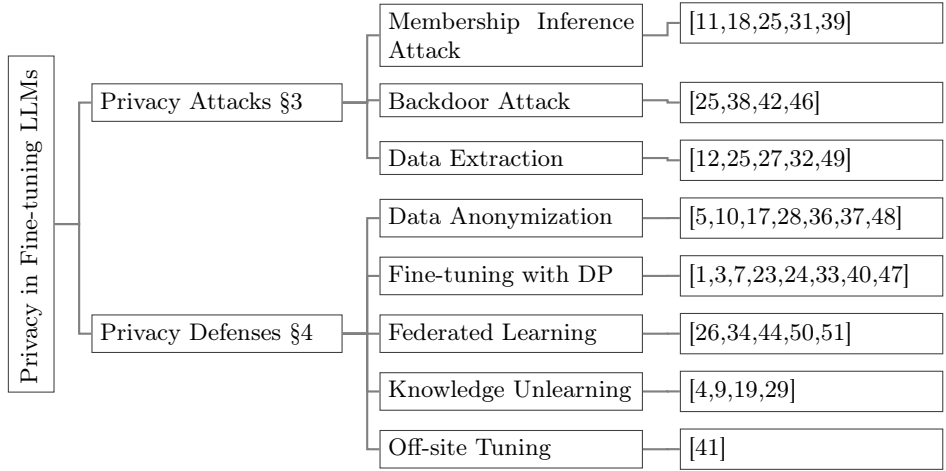
\begin{figure}[t!]
\centering
\tikzset{
    my node/.style={
        font=\small,
        rectangle,
        draw=#1!75,
        align=justify,
    }
}
\forestset{
    my tree style/.style={
        for tree={grow=east,    
            parent anchor=east, %
            child anchor=west,  
        where level=0{my node=black,text width=13em,rotate=90,align=center,anchor=center}{},
        where level=1{my node=black,text width=9.5em}{},
        where level=2{my node=black,text width=10em}{},
        where level=3{my node=black,text width=10em}{},
            l sep=1.5em,
            forked edge,                %
            fork sep=1em,               %
            edge={draw=black!50, thick},                
            if n children=3{for children={
                    if n=2{calign with current}{}}
            }{},
            tier/.option=level,
        }
    }
}
    \begin{forest}
      my tree style
[Privacy in Fine-tuning LLMs
    [Privacy Defenses \S\ref{sec:defenses}
        [Off-site Tuning[\cite{xiao2023offsite}]]
        [Knowledge Unlearning        [\cite{emnlp2023_unlearning,eldan2024whos,jang-etal-2023-knowledge,meng2022locating}]]
        [Federated Learning[\cite{liu2024differentiallyprivatelowrankadaptation,sun2024fedbpt,298559,10447454,emnlp2024_privateFL}]]
        [Fine-tuning with DP [\cite{10031034,10.5555/3618408.3618538,ding2024_differentially_private_transformer,li2022large,li2023privacyprompttuning,emnlp2022_JFT,wu-etal-2022-adaptive,yu2022differentially}]]
        [Data Anonymization [\cite{chen2023hide,8970912,igamberdiev2023dp,7743936,tong2024inferdptprivacypreservinginferenceblackbox,vats2024recovering,ACL21/YueDu21}]]
    ] 
    [Privacy Attacks \S\ref{sec:attacks}
        [Data Extraction[\cite{gupta2022recovering,liu2024precurious,lukas2023analyzing,ozdayi2023controlling,zanella2020analyzing}]]
        [Backdoor Attack[\cite{liu2024precurious,wan2023poisoning,xu2023instructions,yan2024backdooring}]]
        [Membership Inference Attack[\cite{fu2023practical,jagannatha2021membership,liu2024precurious,mireshghallah2022empirical,wen2024privacy}]]
    ]
]
\end{forest}
\centering
\caption{Overview of Privacy in Fine-tuning LLMs.}
\label{fig:Taxonomy of LLMs}
\vspace{-1em}
\end{figure}

Researchers have proposed various privacy attacks that specifically target the fine-tuning stage, exploiting the vulnerabilities unique to this phase. 
Correspondingly, a range of novel defense mechanisms has also been introduced, aiming to mitigate these risks and safeguard sensitive data during fine-tuning.
Those studies~\cite{liu2024precurious,wen2024privacy,jagannatha2021membership,mireshghallah2022empirical} reveal distinct challenges due to two critical factors: (1) the diversity of fine-tuning methods, including various PEFT methods and full fine-tuning; (2) the wide availability of various pre-trained models used as starting points. 
These elements significantly influence privacy risks and defense mechanisms, making fine-tuning a uniquely vulnerable phase.

Existing works~\cite{huang2024harmful,li2023privacy,miranda2024preserving,yan2024protecting} have already attempted to summarize the privacy risks in LLMs and the corresponding defense mechanisms. 
Papers~\cite{li2023privacy,miranda2024preserving,yan2024protecting} on privacy risks and defenses for LLMs tend to adopt a broad perspective, addressing privacy issues across multiple stages of the LLM lifecycle, including pre-training, fine-tuning, and inference. 
While these surveys provide valuable insights, they lack the depth needed to capture the unique aspects of the fine-tuning stage. 
In contrast, the work by Huang et al.~\cite{huang2024harmful}, focuses on attacks and defenses during the fine-tuning stage, but places greater emphasis on model safety rather than privacy.
In conclusion, all of these surveys failed to focus on the two critical factors we mentioned above, including fine-tuning methods and pre-trained models in the fine-tuning phase.
Therefore, it is a obvious lack of studies that offer a comprehensive survey of privacy-preserving techniques in fine-tuning LLMs.

To bridge this gap, our survey focuses exclusively on the fine-tuning stage, offering a deeper exploration of the privacy challenges, attack methodologies, and defense mechanisms relevant to this critical phase, as summarized in Fig. \ref{fig:Taxonomy of LLMs}. 
Specifically, we focused on and summarized the fine-tuning methods and pre-trained models employed in each type of attack and defense.
By comparing and contrasting fine-tuning-specific attacks and defenses, we aim to identify gaps in the existing literature and provide actionable insights to guide future research.
In this survey, we begin by introducing essential background knowledge about fine-tuning LLMs, including an overview of fine-tuning and existing fine-tuning methods for LLMs to provide readers with a foundational understanding of this stage. 
Next, we summarize the current privacy attacks targeting the fine-tuning process, aiming to highlight the unique privacy risks associated with this phase. We then outline the defense mechanisms applicable to the fine-tuning stage, discussing their strengths and limitations. Finally, in the \textit{Challenges and Future Directions} section, we identify research gaps and challenges in this domain and suggest potential directions for future exploration.

\section{Preliminaries}
\label{sec:fine-tuning LLms}
\subsection{Fine-tuning Model}
Fine-tuning is a fundamental process for adapting pre-trained models to downstream tasks, enabling them to leverage the vast knowledge captured during pre-training. 

Fine-tuning involves training a pre-trained model on a smaller, task-specific dataset, which is closely aligned with the target objectives. The datasets used for fine-tuning often come from specialized sources, such as forums, professional communities, or curated data repositories in fields like medicine, law, and technology.
Typically, fine-tuning datasets are composed of labeled demonstration data, including human-labeled datasets, machine-generated datasets, or domain-specific annotated examples. These datasets tend to be relatively small, ranging from a few hundred to several thousand samples, especially when compared to the massive datasets used in the pre-training phase. Through this targeted process, fine-tuning enables the model to learn nuanced characteristics and details pertinent to specific tasks, thereby enhancing its applicability and effectiveness.

Fine-tuning heavily relies on pre-trained models, which are often openly released by other individuals or organizations. These pre-trained models provide a foundational knowledge base that fine-tuning adapts for more specific use cases. In comparison to pre-training, fine-tuning typically involves more sensitive data and leverages task-specific strategies to optimize speed and reduce costs. The selection of fine-tuning methods often depends on the requirements of the target task, highlighting the flexibility and practicality of this process.
\subsection{Parameter-Efficient Fine-Tuning Methods}
While traditional full fine-tuning updates all parameters of a model, achieving high task-specific performance, it is computationally expensive and prone to overfitting, especially with limited data. 
To overcome these challenges, a variety of Parameter-Efficient Fine-Tuning (PEFT) methods \cite{devlin2018bert,ben-zaken-etal-2022-bitfit,lester-etal-2021-power,li-liang-2021-prefix,xu2023parameter,hu2022lora} have been proposed, offering more efficient alternatives. 
These methods vary in their approach, striking different balances between parameter efficiency, adaptability, performance, and speed, expand the range of available fine-tuning methods.

Full Fine-tuning~\cite{devlin2018bert} (FFT) updates all parameters of a model, providing maximum flexibility and allows the model to fully adapt to the task. However, it is computationally intensive and requires substantial data to avoid overfitting, making it less practical for resource-constrained scenarios. 
{\bfseries BitFit}~\cite{ben-zaken-etal-2022-bitfit}, which is Bias-term Fine-tuning, only updates bias terms while freezing the rest of the model. It is lightweight, with minimal computational cost but limited in capturing complex task-specific features.
{\bfseries Prompt-tuning}~\cite{lester-etal-2021-power} introduces trainable prompt vectors to guide the pre-trained model without modifying its internal weights. Prompt-tuning is effective for small datasets, and leverages the full capacity of large pre-trained models but less effective for smaller models.
{\bfseries Prefix-tuning}~\cite{li-liang-2021-prefix} adds trainable prefix vectors at every Transformer layer, interacting with input through attention mechanisms. Compared to Prompt-tuning, it allows dynamic adjustment of intermediate representations, improving adaptability for complex tasks, but slightly increases computational cost due to its per-layer modifications.
{\bfseries Adapters}~\cite{houlsby2019parameter} inserts small task-specific trainable modules into Transformer layers while freezing the backbone. Adapters are well-suited for multi-task learning and are modular by design, enabling efficient transfer across tasks. However, introducing extra modules leads to extra computation, which may impact inference speed.
{\bfseries LoRA}~\cite{hu2022lora} uses low-rank decomposition of weight matrices and fine-tunes only the low-rank components. It maintains a effective balance between parameter efficiency and task performance, making it highly applicable in large-scale models and complex tasks.

Due to the extensive variety of fine-tuning methods for LLMs, this survey focuses on the representative fine-tuning methods introduced above. In the following tables, we use the following abbreviations: A for Adapter, L for LoRA, B for BitFit, P1 for Prompt-tuning, P2 for Prefix-tuning, and F for Full fine-tuning.
\renewcommand{\arraystretch}{1.2}
\begin{table}[h!]
\centering
\caption{Privacy Attacks on Fine-tuning LLMs}
\label{tab:attack_comparison}
\setlength{\tabcolsep}{1.6mm}{
\begin{tabular}{|c|c|c|c|c|c|c|}
\hline
\textbf{Attack} & \textbf{Paper} & \textbf{Target} & \textbf{Capability} & \textbf{FT} & \textbf{Pre-train} & \textbf{Access} \\ \hline
\multirow{10}{*}{MIA} & \cite{mireshghallah2022empirical} & FD & Reference Model & F,A & GPT2 & BB \\ \cline{2-7}
& \cite{jagannatha2021membership} & FD & \diagbox[width=5em,height=1em]{}{} & F & GPT2 & BB,WB \\ \cline{2-7}
& \cite{fu2023practical} & FD & Fine-tuning Access & L & \begin{tabular}[c]{@{}c@{}}GPT2\\ Llama\\ GPTJ\\ Falcon\end{tabular} & BB \\ \cline{2-7}
& \cite{wen2024privacy} & FD & \begin{tabular}[c]{@{}c@{}}Poison Injection\\ while\\ Pre-training\end{tabular} & F,L & \begin{tabular}[c]{@{}c@{}}GPT-Neo\\ Pythia\\ ClinicalBERT\end{tabular} & BB \\ \cline{2-7}
& \cite{liu2024precurious} & FD & Publishing Models & A,B,L & GPT2 & BB \\ \hline
\multirow{7}{*}{DE} & \cite{lukas2023analyzing} & FD & \begin{tabular}[c]{@{}c@{}}Masked\\ Fine-tuning Data\end{tabular} & F & GPT2 & BB \\ \cline{2-7}
& \cite{ozdayi2023controlling} & FD & Fine-tuning Access & P1 & \begin{tabular}[c]{@{}c@{}}GPT2\\ GPT-Neo\end{tabular} & BB \\ \cline{2-7}
& \cite{zanella2020analyzing} & FD & Reference Model & F & \begin{tabular}[c]{@{}c@{}}RNN\\ BERT\end{tabular} & BB \\ \cline{2-7}
& \cite{gupta2022recovering} & FD & \diagbox[width=5em,height=1em]{}{} & F & GPT2 & WB \\ \cline{2-7}
& \cite{liu2024precurious} & FD & Publishing Models & A,B,L & GPT2 & BB \\ \hline
\multirow{7}{*}{BA} & \cite{wen2024privacy} & FD & \begin{tabular}[c]{@{}c@{}}Poison Injection\\ while\\ Pre-training\end{tabular} & F,L & \begin{tabular}[c]{@{}c@{}}GPT-Neo\\ Pythia\\ ClinicalBERT\end{tabular} & BB \\ \cline{2-7}
& \cite{yan2024backdooring} & M & Fine-tuning Access & F & Alpaca & BB \\ \cline{2-7}
& \cite{xu2023instructions} & M & Fine-tuning Access & F,L & \begin{tabular}[c]{@{}c@{}}T5\\ GPT2\\ Llama2\end{tabular} & BB \\ \cline{2-7}
& \cite{wan2023poisoning} & M & Fine-tuning Access & F & T5 & BB \\ \cline{2-7}
& \cite{liu2024precurious} & FD & Publishing Models & A,B,L & GPT2 & BB \\ \hline
\end{tabular}
\begin{tablenotes}
\item[1] MIA: Membership Inference Attack \quad BA: Backdoor Attack \quad DE: Data Extraction
\item[1] FD: Fine-tuning data \quad M: manipulating model \quad F: Full Fine-tuning\quad A: Adapter\quad L: LoRA \quad B: BitFit \quad P1: Prompt-tuning\quad BB: Black-box \quad WB: White-box
\item[1]
\end{tablenotes}
}
\vspace{-3em}
\end{table}

\section{Privacy Attacks in Fine-tuning LLMs}
\label{sec:attacks}
Fine-tuning is a process of adapting pre-trained models using specific datasets to make them more suitable for specialized downstream tasks. However, this process introduces unique privacy risks.
This section focuses on privacy attacks specific to the fine-tuning phase, highlighting two major inherent risks:

\begin{itemize}
    \item Attacks targeting sensitive data in fine-tuning datasets:
These attacks, including membership inference attacks and data extraction attacks, aim to extract sensitive information from fine-tuning data. For example, an adversary may infer whether a specific data point was included in the fine-tuning dataset or even reconstruct the data itself, thereby compromising data confidentiality and integrity.
   \item Attacks aiming at manipulating model outputs:
Often achieved through backdoor attacks, these involve adversaries embedding malicious triggers during the fine-tuning process to control the model's behavior under specific conditions. These attacks can be exploited to induce the model to output sensitive information, and they can be combined with other attack methods to further enhance the overall success rate.
\end{itemize}

By analyzing these targeted threats, this section aims to uncover the unique privacy challenges and attack surfaces associated with fine-tuning, providing a foundation for developing effective defense and mitigation strategies.
\autoref{tab:attack_comparison} compares the attack methods across five dimensions: \textbf{Target}, which specifies the objective of the attack; \textbf{Capability}, which describes the attacker's additional abilities beyond model access; \textbf{FT}, representing the fine-tuning method used in the studies; \textbf{Pre-train}, indicating the pre-trained model employed; \textbf{Access}, referring to the attacker's level of access to the fine-tuned model.

Furthermore, we intend to provide a more detailed explanation of the content within the \textbf{Capability}. \textbf{Reference Model} means the attacker needs the original model or a fine-tuned one as a reference in the attack. \textbf{Fine-tuning Access} 
indicates that an attacker needs to have the access to fine-tune the model. \textbf{Poison Injection while Pre-training} indicates that an attacker has the capability to inject malicious data during the model's pre-training phase. \textbf{Publishing Models} refers to the capability of publicly releasing and sharing models. \textbf{Masked
Fine-tuning Data} means an attacker can get fine-tuning datasets with the sensitive information masked.
\subsection{Membership Inference Attack}
\label{subsec:MIA}
Membership inference attacks (MIAs) aim to identify whether specific data points were part of a model's training dataset, posing significant risks to sensitive information such as clinical records and preference datasets. Recent studies have examined the mechanisms, vulnerabilities, and impacts of MIAs on large language models (LLMs), shedding light on various attack strategies and the factors that influence model susceptibility. Among these, black-box access is the most commonly assumed scenario for MIAs, wherein attackers lack access to the model's internal parameters but infer the inclusion of data samples in the fine-tuning process based on the model's outputs.

Jagannatha et al.~\cite{jagannatha2021membership} conducted a black-box MIA on clinical language models, employing a Threshold-Based Attack to identify which samples were included in fine-tuning across models of varying sizes. 
Their findings revealed that model architecture and size significantly influence privacy risks. 
Specifically, Auto-Regressive Models (ARMs) like GPT-2 are more susceptible to MIAs compared to Masked Language Models (MLMs) such as BERT. 
Moreover, the success rate of these attacks increases with model size.

Mireshghallah et al.~\cite{mireshghallah2022empirical} further leveraged the fine-tuning scenario where attackers can access the pre-trained base model. 
They proposed an MIA based on Likelihood Ratios, using the original model as a reference. 
This approach was extended from masked language models to auto-regressive language models.
Their study demonstrated that the fine-tuning method impacts privacy vulnerability, with head fine-tuning being more vulnerable to MIA compared to full fine-tuning or adapter fine-tuning.
The easy accessibility of pre-trained models exacerbates the risks associated with MIAs. 
Fu et al.~\cite{fu2023practical} introduced SPV-MIA, a novel technique that eliminates the need for attackers to access an external reference dataset. 
This approach uses a self-prompting method to generate a reference dataset directly from the target model. 
By incorporating a probabilistic variation metric, SPV-MIA focuses on measuring memorization rather than overfitting, offering a more robust indicator for identifying member records.

When an attacker can modify the pre-trained model and publish it, the backdoor can be easily integrated into MIA, leading to a significant increase in attack success rate of MIA.
Wen et al.~\cite{wen2024privacy} demonstrated that injecting a backdoor into a pre-trained model before fine-tuning substantially enhances the effectiveness of MIAs. 
Similarly, Liu et al.~\cite{liu2024precurious} proposed the PreCurious attack framework, where the attacker fine-tunes a legitimate pre-trained model using auxiliary data to create an adversarially initialized model. 
Once the victim fine-tunes this model, the attacker can execute highly effective MIAs on the fine-tuned model.

\begin{remark}
Membership inference attacks exploit various characteristics of the fine-tuning process, with factors such as model architecture, size, and fine-tuning methods significantly influencing vulnerability. While traditional MIAs remain effective, emerging approaches introduce new attack vectors. Additionally, the integration of backdoors models further amplifies the risk, underscoring the pressing need for robust defenses tailored to fine-tuning scenarios.
\end{remark}

\subsection{Data Extraction Attack}
\label{subsec:DEA}
Data extraction and reconstruction attacks expose a significant vulnerability in large language models (LLMs), allowing attackers to retrieve sensitive information, including personally identifiable information (PII), from model outputs or internal representations. These attacks highlight the critical need to assess the security implications of deploying LLMs in applications that handle sensitive or proprietary data, such as healthcare, finance, and customer service. Recent studies have investigated various data extraction techniques, demonstrating the increasing sophistication and effectiveness of these attacks, driven by advances in adversarial methods and model interpretability tools. Furthermore, the diversity of attack methodologies, including querying-based approaches and gradient-based approaches, underscores the multifaceted challenges in mitigating these risks for privacy preserving.

Lukas et al.~\cite{lukas2023analyzing} introduced three types of black-box data extraction attacks: PII Extraction, PII Reconstruction, and PII Inference. 
Their key innovation lies in PII reconstruction. Because of the unique characteristics of autoregressive models, previous reconstruction attacks struggled to fully utilize contextual information. To address this limitation, they introduced a novel approach using a pre-trained masked language model to generate potential candidates, significantly improving the success rate of reconstructing masked PII.
Their evaluation also revealed that Differential Privacy (DP) mechanisms reduce the attack's effectiveness but do not fully eliminate the risk.

As previously mentioned, fine-tuning introduces new attack scenarios. 
For example, when attackers gain access to publicly released fine-tuned models and can perform further training or fine-tuning, new vulnerabilities arise. 
Using snapshots of the base model with cumulative change in the probabilities assigned to each token by the two models, Zanella-Béguelin et al.~\cite{zanella2020analyzing} have successfully extracted the data used to fine-tune and update the model.
Ozdayi et al.~\cite{ozdayi2023controlling} demonstrated that through prompt-tuning, attackers can deliberately control the memorization of specific data by the model, enabling sensitive information from the fine-tuning dataset to be more easily extracted. 
PreCurious, which adds a backdoor to the pre-trained models proposed by Liu et al.~\cite{liu2024precurious} also supports data extraction.

Another avenue for data extraction involves gradient-based attacks, often referred to as inversion attacks. 
Gupta et al.~\cite{gupta2022recovering} explored the fine-tuning scenario in a Federated Learning setting and introduced the FILM attack. 
By capturing gradient updates during federated training, FILM leverages the language priors encoded in pre-trained models to recover sentences. 
This approach highlights how gradients can reveal sensitive information, posing additional privacy risks in distributed learning environments.

\begin{remark}
    The easy accessibility of pre-trained models has enabled adversaries to significantly enhance extraction success rates. In particular, the availability of publicly released base models provides attackers with a reliable reference point, creating nearly unavoidable risks. This poses substantial challenges for designing effective defense mechanisms.
\end{remark}

\subsection{Backdoor Attack}
\label{subsec:backdoor}

Backdoor attacks aim to inject malicious behaviors or vulnerabilities into models by manipulating the fine-tuning process, often with minimal effort. 
As discussed in the previous sections, backdoor attacks can be seamlessly integrated into various attack workflows. 
While most backdoor techniques are not explicitly designed for privacy attacks but rather to manipulate model outputs, their privacy risks are nonetheless significant. 
Moreover, since the base models used in fine-tuning are often derived from publicly available, open-source releases, attackers have ample opportunity to introduce malicious backdoors into these models.
In this section, we review some of the latest backdoor attacks, focusing on how backdoors are inserted during fine-tuning. 
These methods typically involve modifications or fine-tuning of publicly available models to implant harmful backdoors, exposing fine-tuned models to increased privacy risks.

Numerous studies have highlighted the vulnerability of instruction-tuned models to backdoor attacks. Wan et al.~\cite{wan2023poisoning} demonstrated that instruction-tuned models, such as T5, are particularly susceptible to such attacks.
By injecting carefully crafted malicious examples into the fine-tuning dataset, attackers can implant backdoors, causing the fine-tuned model to exhibit incorrect behavior in classification tasks when specific trigger phrases are encountered.
Similarly, Xu et al.~\cite{xu2023instructions} found that backdoors can achieve high success rates by mixing malicious instructions with legitimate data.
To investigate this vulnerability further, Yan et al.~\cite{yan2024backdooring} proposed the Virtual Prompt Injection (VPI) attack. By poisoning instruction-tuning data, attackers can precisely control the model's output in response to specific triggers, thereby exposing more pronounced privacy risks.

In addition to manipulating model behavior, backdoor attacks have also been applied to the extraction of private information.
For instance, PreCurious~\cite{liu2024precurious} pre-tuning the model on auxiliary data to enable rapid overfitting on private data during fine-tuning and adjusts model parameters to delay convergence, forcing prolonged exposure to private data and increasing memorization risks. Wen et al.~\cite{wen2024privacy} also proved the a pre-tained model with a backdoor poses a greater risk to the privacy of fine-tuning data.
This backdoor design amplifies the original privacy risks by several magnitudes.

\begin{remark}
    Traditional backdoors can be implanted through fine-tuning to manipulate models into generating harmful outputs, which is particularly effective in instruction tuning. Additionally, backdoor attacks can serve as auxiliary methods to significantly enhance the success rates of MIAs and data extractions. The combination of multiple attack strategies in the context of fine-tuning is relatively straightforward to implement, further amplifying the associated risks.
\end{remark}

\renewcommand{\arraystretch}{1.2} %
\begin{table}[h!]
\centering
\caption{Privacy Defenses in Fine-tuning LLMs}
\begin{tabular}{|cc|c|c|c|}
\hline
\multicolumn{2}{|c|}{\textbf{Defense}} & \textbf{Paper} & \textbf{Pre-train} & \textbf{Fine-tuning} \\ \hline
\multicolumn{1}{|c|}{\multirow{3}{*}{\begin{tabular}[c]{@{}c@{}}Data\\ Anonymization\end{tabular}}} & \begin{tabular}[c]{@{}c@{}}Classical\\ Anonymization\end{tabular} & \cite{7743936},\cite{vats2024recovering},\cite{chen2023hide} & {\multirow{3}{*}{\begin{tabular}[c]{@{}c@{}}\diagbox[width=4em,height=3em]{}{}\end{tabular}}} & {\multirow{3}{*}{\begin{tabular}[c]{@{}c@{}}\diagbox[width=4em,height=3em]{}{}\end{tabular}}} \\ \cline{2-3}
\multicolumn{1}{|c|}{} & \begin{tabular}[c]{@{}c@{}}Anonymization\\ with DP\end{tabular} & \cite{8970912},\cite{igamberdiev2023dp},\cite{ACL21/YueDu21},\cite{tong2024inferdptprivacypreservinginferenceblackbox} & & \\ \hline
\multicolumn{1}{|c|}{\multirow{9}{*}{\begin{tabular}[c]{@{}c@{}}Fine-tuning\\ with DP\end{tabular}}} & \multirow{6}{*}{\begin{tabular}[c]{@{}c@{}}Full\\ Fine-tuning\\ with DP\end{tabular}} & \cite{li2022large} & \begin{tabular}[c]{@{}c@{}}BERT,RoBERTa,\\ GPT2\end{tabular} & \multirow{6}{*}{F} \\ \cline{3-4}
\multicolumn{1}{|c|}{} & & \cite{ding2024_differentially_private_transformer} & BERT,GPT2 & \\ \cline{3-4}
\multicolumn{1}{|c|}{} & & \cite{10.5555/3618408.3618538} & RoBERTa,GPT2 & \\ \cline{3-4}
\multicolumn{1}{|c|}{} & & \cite{10031034} & RoBERTa & \\ \cline{3-4}
\multicolumn{1}{|c|}{} & & \cite{emnlp2022_JFT} & RoBERTa,GPT2 & \\ \cline{3-4}
\multicolumn{1}{|c|}{} & & \cite{wu-etal-2022-adaptive} & Custom TF & \\ \cline{2-5}
\multicolumn{1}{|c|}{} & \multirow{3}{*}{\begin{tabular}[c]{@{}c@{}}PEFT\\ with DP\end{tabular}} & \cite{yu2022differentially} & GPT2,RoBERTa & L,A \\ \cline{3-5}
\multicolumn{1}{|c|}{} & & \cite{li2022large} & GPT2,RoBERTa & L,P2 \\ \cline{3-5}
\multicolumn{1}{|c|}{} & & \cite{li2023privacyprompttuning} & BERT & P1,P2 \\ \hline
\multicolumn{2}{|c|}{\multirow{5}{*}{Federated Learning}} & \cite{298559} & \begin{tabular}[c]{@{}c@{}}ALBERT,BERT,\\ DistilBERT,RoBERTa\end{tabular} & A,L,B \\ \cline{3-5}
\multicolumn{2}{|c|}{} & \cite{10447454} & Llama & L \\ \cline{3-5}
\multicolumn{2}{|c|}{} & \cite{sun2024fedbpt} & RoBERTa,Llama2 & P1 \\ \cline{3-5}
\multicolumn{2}{|c|}{} & \cite{emnlp2024_privateFL} & Llama2,ChatGLM & P1 \\ \cline{3-5}
\multicolumn{2}{|c|}{} & \cite{liu2024differentiallyprivatelowrankadaptation} & Llama,ChatGLM & L \\ \hline
\multicolumn{2}{|c|}{\multirow{4}{*}{Knowledge Unlearning}} & \cite{emnlp2023_unlearning} & T5 & F \\ \cline{3-5}
\multicolumn{2}{|c|}{} & \cite{meng2022locating} & GPT-NEO,OPT& F\\ \cline{3-5}
\multicolumn{2}{|c|}{} & \cite{jang-etal-2023-knowledge} & GPT-Neo,OPT &F \\ \cline{3-5}
\multicolumn{2}{|c|}{} & \cite{eldan2024whos} & Llama2 & F \\ \hline
\multicolumn{2}{|c|}{Off-site tuning} & \cite{xiao2023offsite} & \begin{tabular}[c]{@{}c@{}}GPT2,OPT,\\ BLOOM\end{tabular} & A,L,B \\ \hline
\end{tabular}
\begin{tablenotes}
     \item[1]  F: Full Fine-tuning\quad A: Adapter\quad L: LoRA \quad B: BitFit \quad \\ P1: Prompt-tuning\quad  P2: Prefix-tuning
   \end{tablenotes}
\label{tab:defense}
\vspace{-1em}
\end{table}

\section{Privacy Defenses in Fine-tuning LLMs}
\label{sec:defenses}
In this section, we investigate existing privacy defense methods that can be used in fine-tuning phase. 
\autoref{tab:defense} compares the common privacy defense methods: data anonymization, fine-tuning with DP, federated learning, knowledge unlearning, and off-site tuning.

\subsection{Data Anonymization}

Data anonymization protects privacy data security by removing or obfuscating sensitive information and employing techniques such as data masking, aggregation, generalization, or randomization to reduce the identifiability of sensitive data. 
The k-anonymity~\cite{sweeney1997guaranteeing} was firstly introduced in 1997, which means ensuring that data cannot be re-identified to fewer than k individuals. This fundamental concept has driven significant research, resulting in the development of numerous anonymization techniques. In our survey, data anonymization is the process of protecting sensitive information in a text by removing or modifying personally identifiable information (PII) or other confidential details. Since this survey focuses on the fine-tuning stage, and data anonymization can be applied in both pre-training and fine-tuning without being specific to either stage, only a selection of representative methods is introduced in this paper.
\subsubsection{Classical Anonymization.}
The standard classical anonymization workflow involves preprocessing the text, using Named Entity Recognition (NER) to identify sensitive information, applying Coreference Resolution (CRR) to maintain consistency across references, and finally anonymizing the data through methods like suppression, tagging, random substitution, or generalization. 

Mamede et al.~\cite{7743936} developed a four-module anonymization pipeline for Portuguese which closely resembles the general workflow for this task mentioned before. Their work demonstrates the importance of NER and CRR in this workflow which lays a foundation for the future.
More recent anonymization methods leverage the power of LLMs. The LLMs are also used in the classical anonymization workflows by Vats et al.~\cite{vats2024recovering}
Their framework find named entities with NER and mask them. 
After that, an LLM is used to impute previously masked tokens. 
Chen et al.\cite{chen2023hide} further advanced the utilization of LLMs. 
They employed two local LLMs: the first anonymizes the prompt before sending it to a black-box LLM, while the second restores the original content by de-anonymizing the output from the black-box LLM. 
Both of the utilization of LLMs achieve good privacy-utility trade-off.
\subsubsection{Anonymization with Differential Privacy.}
Differential Privacy (DP)~\cite{dwork2006differential} is a formal framework for privacy preservation that ensures the output of an algorithm remains statistically indistinguishable, whether or not a specific individual's data is included in the input dataset.
The adoption of DP in text anonymization offers robust privacy guarantees. Typically, A neural model is utilized to generate embeddings from the input data. These embeddings, which are often clipped, are then subjected to perturbation by adding noise, following the principles of DP, to ensure privacy preservation. Finally, the perturbed embeddings are decoded back into anonymized output, balancing utility and privacy protection.
Feyisetan et al.~\cite{8970912} proposed method that leverages hierarchical word embeddings in hyperbolic space to achieve $d_\chi$-privacy for text anonymization with new noise distribution specifically designed for DP. Their work demonstrates an advantageous balance between privacy and utility.

Transformer-based models are proven to be suitable for this kind of DP adoption. DP-BART, introduced by Igamberdiev and Ivan Habernal~\cite{igamberdiev2023dp}, a privatization model based on the BART architecture. Similar to prior approaches using LSTMs, it applies clipping and noise addition to the output of the latent encoder.
However, Yue et al.~\cite{ACL21/YueDu21} argued that directly injecting noise into large, high-dimensional word embeddings disrupts the privacy-utility balance. To address this, they introduced SANTEXT, where words are first mapped into embeddings, distances are calculated, and replacements are sampled with probabilities inversely proportional to these distances. Based on their work, Tong et al.~\cite{tong2024inferdptprivacypreservinginferenceblackbox} use random adjacency lists to increase the defense of possible embedding inversion attacks. By using black-box LLMs to generate open-ended text, their method, RANTEXT, shows very strong performance.
\subsection{Fine-tuning with Differential Privacy}
Differential Privacy(DP) was introduced to model training through Differentially Private Stochastic Gradient Descent (DP-SGD), which is a modification of the standard SGD algorithm designed to ensure differential privacy during training. 
However, applying DP learning to large NLP models has faced significant challenges, including substantial performance degradation and high computational overhead when using DP-SGD, prompting researchers to propose more advanced DP-tuning methods.
\subsubsection{Full Fine-tuning with DP.}
Li et al.~\cite{8970912}addressed these issues by leveraging large pre-trained models, optimizing hyperparameters for DP, and aligning fine-tuning objectives with pretraining procedures. 
They introduced a memory-saving technique known as ghost clipping, which enables efficient DP-SGD for large Transformers. 
This approach maintains strong performance under the same privacy budget with minimal runtime overhead.

Building on ghost clipping, numerous advancements have been proposed. 
For example, the Book-Keeping (BK)~\cite{10.5555/3618408.3618538} technique requires only a single back-propagation round and eliminates the need to instantiate per-sample gradients. 
This method significantly improves throughput while maintaining similar memory usage, providing a more scalable solution for handling high-dimensional data. 
Similarly, Ding et al.~\cite{ding2024_differentially_private_transformer} identified the attention distraction phenomenon in existing approaches, which affects training efficiency and performance in differential privacy settings. 
To address this, they proposed Phantom Clipping, further improving the training of large language models (LLMs) under DP constraints.

In parallel with these improvements, several frameworks for fine-tuning LLMs with differential privacy guarantees have been introduced. 
EW-Tune~\cite{10031034} is a framework that adopts the Edgeworth Accountant method, providing finite-sample privacy guarantees tailored to fine-tuning scenarios. 
Another framework, Fine-tune Twice (JFT)~\cite{emnlp2022_JFT}, selectively applies differential privacy only to the sensitive portions of the data, thereby significantly preserving the utility of the fine-tuned model. 
Wu et al.~\cite{wu-etal-2022-adaptive} introduced an innovative Adaptive Differential Privacy (ADP) framework, which features a novel Adam algorithm that dynamically adjusts the strength of injected differential privacy noise based on estimated privacy probabilities.
\subsubsection{PEFT with Differential Privacy.}
Modern large language models (LLMs) often contain billions of parameters, making full fine-tuning impractical. 
As a result, Parameter-Efficient Fine-Tuning (PEFT) methods, such as LoRA, have become more widely adopted in real-world fine-tuning scenarios. 
This shift has also led to the emergence of differentially private (DP) versions of these PEFTs.

Yu et al.~\cite{yu2022differentially} applied DP techniques to common PEFT methods like LoRA and Adapters, demonstrating the feasibility of DP-PEFT. 
They further highlighted that LoRA, when combined with DP-SGD, excels in terms of speed and memory efficiency, making it particularly suitable for private fine-tuning.
Li et al.~\cite{li2022large} extended existing findings from full fine-tuning to PEFT by incorporating DP-Adam and the Ghost Clipping technique. 
This approach significantly reduced memory usage, narrowing the efficiency gap between private and non-private training to a negligible level. 
Additionally, the performance degradation was proven to be minimal, emphasizing the potential of DP-PEFT in private fine-tuning applications.

Similar to advancements in training frameworks for full fine-tuning, novel frameworks have also been introduced for DP-PEFT. 
RAPT~\cite{li2023privacyprompttuning}, for instance, is designed for prompt-tuning and prefix-tuning under a local privacy setting, making it possible to privately train a remote model. 
This framework demonstrates how PEFTs can be adapted to ensure privacy without sacrificing practicality.
\subsection{Federated Learning}
Federated Learning (FL) is a distributed machine learning paradigm where multiple devices collaboratively train a shared model without sharing their raw data. 
Instead, each device computes updates locally and only exchanges model parameters or gradients with a central server, preserving data privacy while enabling large-scale model training across decentralized datasets. 

FwdLLM~\cite{298559} introduces federated learning into LLM fine-tuning by proposing a backpropagation-free training method called \textit{perturbed inference}. 
This approach replaces traditional gradient computation, significantly reducing memory and computational demands while seamlessly integrating with PEFT methods. 
FedIT~\cite{10447454} framework, which focuses directly on PEFT by leveraging LoRA for instruction-tuning in a federated learning setup. 
Similarly, the FedBPT~\cite{sun2024fedbpt} framework enables prompt-tuning of pre-trained models within federated learning settings. These works collectively contribute to reducing memory usage and computational overhead.
Zheng et al.~\cite{emnlp2024_privateFL} further advanced the field by proposing the FL-GLM framework, which incorporates gradient protection and key encryption techniques to enhance privacy security. 
Additionally, federated learning can be combined with differential privacy techniques. 
Liu et al.~\cite{liu2024differentiallyprivatelowrankadaptation} introduced an implementation of DP-LoRA in federated learning settings, providing robust privacy guarantees while optimizing communication overhead, making it more practical for real-world applications.
\subsection{Knowledge Unlearning}
Knowledge Unlearning, also known as machine unlearning~\cite{emnlp2023_unlearning}, refers to the process by which a machine learning model is able to forget or remove knowledge about certain data points from its training. 
Unlike Differential Privacy, which cannot fully guarantee the \textit{right to be forgotten} due to the inherent limitation of a non-zero privacy budget, Machine Unlearning focuses explicitly on eliminating the influence of specific data samples. 
In the fine-tuning phase, it can also be applied to remove the impact of specific data points in the fine-tuning datasets.

Meng et al.~\cite{meng2022locating} introduced ROME (Rank-One Model Editing), a method that identifies and individually manipulates layers and neurons responsible for factual predictions. 
While originally designed for factual associations, ROME shows potential for broader applications to other types of data. 
Jang et al.~\cite{jang-etal-2023-knowledge} proposed a method for unlearning specific information in language models by maximizing the training loss on target sequences, achieving unlearning with minimal performance degradation. 
Eldan and Russinovich~\cite{eldan2024whos} highlighted challenges in scenarios like making LLaMA2-7B forget specific topics, such as \textit{Harry Potter}. 
Their solution involved replacing specific terms with generic equivalents and training a reinforced model to reduce token likelihoods, requiring numerous gradient descent steps.
Chen and Yang~\cite{emnlp2023_unlearning} approached unlearning from a structural perspective. 
They created unlearning layers and trained them using a selective student-teacher objective. 
Different unlearning layers were used to forget specific information and could eventually be fused into a single layer. 
\subsection{Off-site Tuning}
Off-site Tuning, detailed by Xiao et al.~\cite{xiao2023offsite}, is a novel framework for privacy preserving and efficient transfer learning, designed to address challenges in tuning large-scale pre-trained models without requiring full access to the model or the data. The key idea of Off-site Tuning is to enable transfer learning without sharing full models or sensitive data. It splits the pre-trained model into two components: the Adapter, a small, trainable module for task-specific tuning, and the Emulator, a compressed version of the model that provides gradients while concealing the complete model structure. The data owner fine-tunes the adapter locally with the emulator and sends back the adapter. The model owner integrates it with the full model to complete the tuning. This approach ensures privacy, reduces computational cost, and maintains performance close to full fine-tuning, showing good potential.
\section{Challenges and Future Directions}
\subsection{Privacy Risks from Pre-trained Models}
During the investigation, we found there is a particular reliance on pre-trained models in fine-tuning phase, which introduces new privacy challenges due to the open availability of many pre-trained models:
Adversaries can release maliciously modified pre-trained models, embedding backdoors or vulnerabilities. Fine-tuning these models can amplify privacy risks and increase the likelihood of sensitive information in the fine-tuning dataset being exposed.
Moreover, since most pre-trained models are publicly available, adversaries can easily acquire and use them as reference points or further fine-tune them to facilitate attacks on fine-tuning datasets, significantly increasing attack success rates.
To counter malicious pre-trained models, we need to focus on developing robust verification methods to detect backdoors or malicious modifications in pre-trained models.
Establishing trusted auditing frameworks, such as watermarking mechanisms, can also help the provenance and integrity of pre-trained models.
Enhancing privacy protection during fine-tuning by obfuscating relationships between pre-trained and fine-tuned models is also a possible direction to reduce the effectiveness of reference model.
\subsection{Risks and Opportunities from PEFTs}
Parameter-Efficient Fine-Tuning (PEFT) methods significantly reduce computational demands and fine-tuning time, but they introduce novel attack and defense challenges at the same time.
Privacy implications of different PEFTs remain underexplored, leaving critical questions unanswered:
How do various fine-tuning methods (e.g., LoRA, Adapters, Prefix-tuning) impact privacy risks?
Does PEFT offer better or worse privacy risks compared to full fine-tuning?
What are the differences in privacy risks across PEFT methods, and what causes them?
From an attack perspective, PEFT introduces new vulnerabilities, such as malicious attacks leveraging shared fine-tuning results (e.g., LoRA).
From a defense perspective, addressing these risks remains challenge. 
While PEFT is being increasingly adopted, there is a lack of corresponding research on privacy-preserving methods specifically designed for PEFT.
Therefore, we need to conduct more in-depth and comprehensive research on PEFT methods to understand their varying impacts on privacy.
We can conduct in-depth researches of PEFT-specific attacks to identify new privacy risks and compare vulnerabilities across different PEFT methods.
As for defense, we need develop targeted defenses for the unique attack surfaces introduced by PEFT.
For the possibility of PEFTs, We can explore privacy-preserving methods specifically designed for PEFT or improve existing frameworks, to fully leverage PEFT's efficiency advantages and achieve a better utility-privacy balance.
\subsection{Unique Characteristics of Fine-tuning Data}
Fine-tuning datasets differ significantly from the general-purpose data used in pre-training, often exhibiting narrower domain coverage, task-specific distributions, and higher data sensitivity. For instance, a fine-tuning dataset built from medical records or legal documents typically contains specialized terminology and sensitive personal information, unlike the vast and heterogeneous corpora used in pre-training. These characteristics introduce unique privacy risks.
Narrower and task-specific distributions make fine-tuning datasets more vulnerable to targeted attacks. 
A fine-tuning dataset for legal document summarization might inadvertently retain verbatim excerpts from confidential case files, which could then be extracted by extraction attacks.
However, there is currently a lack of privacy research specifically focused on the characteristics of fine-tuning data. To address this issue, future research can be conducted from both the attack and defense perspectives. 
From the perspective of attack, exploring domain-specific or task-specific attacks can help us understand the unique traits of fine-tuning datasets and design more effective defenses.
From the perspective of defense, we need to explore task-specific or domain-specific privacy mechanisms to adapt to more specific tasks.
\subsection{Limitations of Existing Defense Methods}
Current defense methods struggle to comprehensively address diverse privacy attacks. For instance, Differential Privacy (DP) reduces but does not eliminate risks of data extraction.~\cite{lukas2023analyzing} Fine-tuning can also be used to enhance the memorization of LLMs~\cite{ozdayi2023controlling}, which may make unlearning ineffective. 
Therefore, there is a lack of unified metrics for evaluating effectiveness of defenses across diverse scenarios.
To overcome these limitations, it is advocated to develop comprehensive privacy metrics that evaluate defenses against diverse attack types in fine-tuning settings.
Furthermore, designing integrated defense frameworks that combine multiple privacy-preserving techniques to address a wider range of attacks is also crucial.
\subsection{Balancing Utility and Privacy in Fine-tuning Defenses}
The utility-privacy trade-off remains a longstanding challenge in fine-tuning large language models (LLMs). While defense methods aim to mitigate privacy risks, preserving model performance for downstream tasks remains equally critical. In fact, most defense methods also consider their utility-privacy trade-off as a critical factor. This trade-off, a well-known challenge in privacy protection, continues to be relevant in the fine-tuning stage. Moreover, given the high performance demands of fine-tuning for downstream tasks, maintaining utility becomes even more critical and cannot be overlooked. To address that, dynamic adjustment defense mechanisms or task-aware defense strategies can be potential directions.

\section{Conclusion}
In this survey, we first introduce essential background knowledge about fine-tuning LLMs, including the overview of fine-tuning and several representative fine-tuning methods.
Subsequently, we investigate the existing privacy attacks and defense mechanisms specific to the fine-tuning stage, with a particular focus on how different fine-tuning methods and pre-trained models are employed in these research settings. 
Through this investigation, we aim to uncover gaps between privacy attacks and defenses in the fine-tuning context. 
Finally, we identify key challenges arising from the unique characteristics and emerging techniques of fine-tuning and proposed potential directions.

\section{Acknowledgment}
This work is partially support by JSPS KAKENHI JP23K24851, JST PRESTO JPMJPR23P5, JST CREST JPMJCR21M2, JSPS KAKENHI Grant Number JP24H00685. Lei Chen’s work is partially supported by National Key Research and Development Program of China Grant No. 2023YFF0725100, National Science Foundation of China (NSFC) under Grant No. U22B2060, Guangdong-Hong Kong Technology Innovation Joint Funding Scheme Project No. 2024-A0505040012, the Hong Kong RGC GRF Project 16213620, RIF Project R6020-19, AOE Project AoE/E-603/18, Theme-based project TRS T41-603/20R, CRF Project C2004-21G, Guangdong Province Science and Technology Plan Project 2023A0505030011, Guangzhou municipality big data intelligence key lab, 2023-A03J0012, Hong Kong ITC ITF grants MHX/078/21 and PRP/004/22FX, Zhujiang scholar program 2021JC02X170, Microsoft Research Asia Collaborative Research Grant, HKUST-Webank joint research lab and 2023 HKUST Shenzhen-Hong Kong Collaborative Innovation Institute Green Sustainability Special Fund, from Shui On Xintiandi and the InnoSpace GBA.
%
%
\newpage
\clearpage
\bibliographystyle{splncs04}
\bibliography{Reference}

\begin{thebibliography}{10}
\providecommand{\url}[1]{\texttt{#1}}
\providecommand{\urlprefix}{URL }
\providecommand{\doi}[1]{https://doi.org/#1}

\bibitem{10031034}
Behnia, R., Ebrahimi, M.R., Pacheco, J., Padmanabhan, B.: Ew-tune: A framework for privately fine-tuning large language models with differential privacy. In: 2022 IEEE International Conference on Data Mining Workshops. pp. 560--566 (2022)

\bibitem{ben-zaken-etal-2022-bitfit}
Ben~Zaken, E., Goldberg, Y., Ravfogel, S.: {B}it{F}it: Simple parameter-efficient fine-tuning for transformer-based masked language-models. In: Proceedings of the 60th Annual Meeting of the Association for Computational Linguistics. pp.~1--9 (2022)

\bibitem{10.5555/3618408.3618538}
Bu, Z., Wang, Y.X., Zha, S., Karypis, G.: Differentially private optimization on large model at small cost. In: Proceedings of the 40th International Conference on Machine Learning (2023)

\bibitem{emnlp2023_unlearning}
Chen, J., Yang, D.: Unlearn what you want to forget: Efficient unlearning for llms. In: Proceedings of the 2023 Conference on Empirical Methods in Natural Language Processing. pp. 12041--12052 (2023)

\bibitem{chen2023hide}
Chen, Y., Li, T., Liu, H., Yu, Y.: Hide and seek (has): A lightweight framework for prompt privacy protection. arXiv preprint arXiv:2309.03057  (2023)

\bibitem{devlin2018bert}
Devlin, J.: Bert: Pre-training of deep bidirectional transformers for language understanding. arXiv preprint arXiv:1810.04805  (2018)

\bibitem{ding2024_differentially_private_transformer}
Ding, Y., Wu, X., Meng, Y., Luo, Y., Wang, H., Pan, W.: Delving into differentially private transformer. In: Proceedings of the 41st International Conference on Machine Learning. vol.~235, p.~TBD (2024)

\bibitem{dwork2006differential}
Dwork, C.: Differential privacy. In: International colloquium on automata, languages, and programming. pp. 1--12 (2006)

\bibitem{eldan2024whos}
Eldan, R., Russinovich, M.: Who's harry potter? approximate unlearning in llms. arXiv preprint arXiv:2310.02238  (2023)

\bibitem{8970912}
Feyisetan, O., Diethe, T., Drake, T.: { Leveraging Hierarchical Representations for Preserving Privacy and Utility in Text }. In: 2019 IEEE International Conference on Data Mining. pp. 210--219 (2019)

\bibitem{fu2023practical}
Fu, W., Wang, H., Gao, C., Liu, G., Li, Y., Jiang, T.: Practical membership inference attacks against fine-tuned large language models via self-prompt calibration. arXiv preprint arXiv:2311.06062  (2023)

\bibitem{gupta2022recovering}
Gupta, S., Huang, Y., Zhong, Z., Gao, T., Li, K., Chen, D.: Recovering private text in federated learning of language models. Advances in neural information processing systems  \textbf{35},  8130--8143 (2022)

\bibitem{han2024parameter}
Han, Z., Gao, C., Liu, J., Zhang, J., Zhang, S.Q.: Parameter-efficient fine-tuning for large models: A comprehensive survey. arXiv preprint arXiv:2403.14608  (2024)

\bibitem{houlsby2019parameter}
Houlsby, N., Giurgiu, A., Jastrzebski, S., Morrone, B., De~Laroussilhe, Q., Gesmundo, A., Attariyan, M., Gelly, S.: Parameter-efficient transfer learning for nlp. In: International conference on machine learning. pp. 2790--2799 (2019)

\bibitem{hu2022lora}
Hu, E.J., yelong shen, Wallis, P., Allen-Zhu, Z., Li, Y., Wang, S., Wang, L., Chen, W.: Lo{RA}: Low-rank adaptation of large language models. In: International Conference on Learning Representations (2022)

\bibitem{huang2024harmful}
Huang, T., Hu, S., Ilhan, F., Tekin, S.F., Liu, L.: Harmful fine-tuning attacks and defenses for large language models: A survey. arXiv preprint arXiv:2409.18169  (2024)

\bibitem{igamberdiev2023dp}
Igamberdiev, T., Habernal, I.: Dp-bart for privatized text rewriting under local differential privacy. In: Findings of the Association for Computational Linguistics: ACL 2023. pp. 13914--13934 (2023)

\bibitem{jagannatha2021membership}
Jagannatha, A., Rawat, B.P.S., Yu, H.: Membership inference attack susceptibility of clinical language models. arXiv preprint arXiv:2104.08305  (2021)

\bibitem{jang-etal-2023-knowledge}
Jang, J., Yoon, D., Yang, S., Cha, S., Lee, M., Logeswaran, L., Seo, M.: Knowledge unlearning for mitigating privacy risks in language models. In: Proceedings of the 61st Annual Meeting of the Association for Computational Linguistics. pp. 14389--14408 (2023)

\bibitem{lester-etal-2021-power}
Lester, B., Al-Rfou, R., Constant, N.: The power of scale for parameter-efficient prompt tuning. In: Proceedings of the 2021 Conference on Empirical Methods in Natural Language Processing. pp. 3045--3059 (2021)

\bibitem{li2023privacy}
Li, H., Chen, Y., Luo, J., Wang, J., Peng, H., Kang, Y., Zhang, X., Hu, Q., Chan, C., Xu, Z., et~al.: Privacy in large language models: Attacks, defenses and future directions. arXiv preprint arXiv:2310.10383  (2023)

\bibitem{li-liang-2021-prefix}
Li, X.L., Liang, P.: Prefix-tuning: Optimizing continuous prompts for generation. In: Proceedings of the 59th Annual Meeting of the Association for Computational Linguistics and the 11th International Joint Conference on Natural Language Processing. pp. 4582--4597 (2021)

\bibitem{li2022large}
Li, X., Tramer, F., Liang, P., Hashimoto, T.: Large language models can be strong differentially private learners. In: International Conference on Learning Representations (2022)

\bibitem{li2023privacyprompttuning}
Li, Y., Tan, Z., Liu, Y.: Privacy-preserving prompt tuning for large language model services. arXiv preprint arXiv:2305.06212  (2023)

\bibitem{liu2024precurious}
Liu, R., Wang, T., Cao, Y., Xiong, L.: Precurious: How innocent pre-trained language models turn into privacy traps. arXiv preprint arXiv:2403.09562  (2024)

\bibitem{liu2024differentiallyprivatelowrankadaptation}
Liu, X.Y., Zhu, R., Zha, D., Gao, J., Zhong, S., White, M., Qiu, M.: Differentially private low-rank adaptation of large language model using federated learning (2024)

\bibitem{lukas2023analyzing}
Lukas, N., Salem, A., Sim, R., Tople, S., Wutschitz, L., Zanella-B{\'e}guelin, S.: Analyzing leakage of personally identifiable information in language models. In: 2023 IEEE Symposium on Security and Privacy. pp. 346--363 (2023)

\bibitem{7743936}
Mamede, N., Baptista, J., Dias, F.: Automated anonymization of text documents. In: 2016 IEEE Congress on Evolutionary Computation. pp. 1287--1294 (2016)

\bibitem{meng2022locating}
Meng, K., Bau, D., Andonian, A., Belinkov, Y.: Locating and editing factual associations in gpt. Advances in Neural Information Processing Systems  \textbf{35},  17359--17372 (2022)

\bibitem{miranda2024preserving}
Miranda, M., Ruzzetti, E.S., Santilli, A., Zanzotto, F.M., Brati{\`e}res, S., Rodol{\`a}, E.: Preserving privacy in large language models: A survey on current threats and solutions. arXiv preprint arXiv:2408.05212  (2024)

\bibitem{mireshghallah2022empirical}
Mireshghallah, F., Uniyal, A., Wang, T., Evans, D.K., Berg-Kirkpatrick, T.: An empirical analysis of memorization in fine-tuned autoregressive language models. In: EMNLP. pp. 1816--1826 (2022)

\bibitem{ozdayi2023controlling}
Ozdayi, M.S., Peris, C., FitzGerald, J., Dupuy, C., Majmudar, J., Khan, H., Parikh, R., Gupta, R.: Controlling the extraction of memorized data from large language models via prompt-tuning. arXiv preprint arXiv:2305.11759  (2023)

\bibitem{emnlp2022_JFT}
Shi, W., Shea, R., Chen, S., Zhang, C., Jia, R., Yu, Z.: Just fine-tune twice: Selective differential privacy for large language models. In: Proceedings of the 2022 Conference on Empirical Methods in Natural Language Processing. pp. 6327--6340 (2022)

\bibitem{sun2024fedbpt}
Sun, J., Xu, Z., Yin, H., Yang, D., Xu, D., Chen, Y., Roth, H.R.: Fed{BPT}: Efficient federated black-box prompt tuning for large language models (2024)

\bibitem{sweeney1997guaranteeing}
Sweeney, L.: Guaranteeing anonymity when sharing medical data, the datafly system. In: Proceedings of the AMIA Annual Fall Symposium. p.~51 (1997)

\bibitem{tong2024inferdptprivacypreservinginferenceblackbox}
Tong, M., Chen, K., Zhang, J., Qi, Y., Zhang, W., Yu, N., Zhang, T., Zhang, Z.: Inferdpt: Privacy-preserving inference for black-box large language model (2024)

\bibitem{vats2024recovering}
Vats, A., Liu, Z., Su, P., Paul, D., Ma, Y., Pang, Y., Ahmed, Z., Kalinli, O.: Recovering from privacy-preserving masking with large language models. In: ICASSP 2024-2024 IEEE International Conference on Acoustics, Speech and Signal Processing. pp. 10771--10775 (2024)

\bibitem{wan2023poisoning}
Wan, A., Wallace, E., Shen, S., Klein, D.: Poisoning language models during instruction tuning. In: International Conference on Machine Learning. pp. 35413--35425 (2023)

\bibitem{wen2024privacy}
Wen, Y., Marchyok, L., Hong, S., Geiping, J., Goldstein, T., Carlini, N.: Privacy backdoors: Enhancing membership inference through poisoning pre-trained models. arXiv preprint arXiv:2404.01231  (2024)

\bibitem{wu-etal-2022-adaptive}
Wu, X., Gong, L., Xiong, D.: Adaptive differential privacy for language model training. In: Proceedings of the First Workshop on Federated Learning for Natural Language Processing. pp. 21--26 (2022)

\bibitem{xiao2023offsite}
Xiao, G., Lin, J., Han, S.: Offsite-tuning: Transfer learning without full model. arXiv  (2023)

\bibitem{xu2023instructions}
Xu, J., Ma, M.D., Wang, F., Xiao, C., Chen, M.: Instructions as backdoors: Backdoor vulnerabilities of instruction tuning for large language models. arXiv preprint arXiv:2305.14710  (2023)

\bibitem{xu2023parameter}
Xu, L., Xie, H., Qin, S.Z.J., Tao, X., Wang, F.L.: Parameter-efficient fine-tuning methods for pretrained language models: A critical review and assessment. arXiv preprint arXiv:2312.12148  (2023)

\bibitem{298559}
Xu, M., Cai, D., Wu, Y., Li, X., Wang, S.: {FwdLLM}: Efficient federated finetuning of large language models with perturbed inferences. In: 2024 USENIX Annual Technical Conference. pp. 579--596 (2024)

\bibitem{yan2024protecting}
Yan, B., Li, K., Xu, M., Dong, Y., Zhang, Y., Ren, Z., Cheng, X.: On protecting the data privacy of large language models (llms): A survey. arXiv preprint arXiv:2403.05156  (2024)

\bibitem{yan2024backdooring}
Yan, J., Yadav, V., Li, S., Chen, L., Tang, Z., Wang, H., Srinivasan, V., Ren, X., Jin, H.: Backdooring instruction-tuned large language models with virtual prompt injection. In: Proceedings of the 2024 Conference of the North American Chapter of the Association for Computational Linguistics: Human Language Technologies. pp. 6065--6086 (2024)

\bibitem{yu2022differentially}
Yu, D., Naik, S., Backurs, A., Gopi, S., Inan, H.A., Kamath, G., Kulkarni, J., Lee, Y.T., Manoel, A., Wutschitz, L., Yekhanin, S., Zhang, H.: Differentially private fine-tuning of language models. In: International Conference on Learning Representations (2022)

\bibitem{ACL21/YueDu21}
Yue, X., Du, M., Wang, T., Li, Y., Sun, H., Chow, S.S.M.: Differential privacy for text analytics via natural text sanitization. In: Findings, {ACL-IJCNLP} 2021 (2021)

\bibitem{zanella2020analyzing}
Zanella-B{\'e}guelin, S., Wutschitz, L., Tople, S., R{\"u}hle, V., Paverd, A., Ohrimenko, O., K{\"o}pf, B., Brockschmidt, M.: Analyzing information leakage of updates to natural language models. In: Proceedings of the 2020 ACM SIGSAC conference on computer and communications security. pp. 363--375 (2020)

\bibitem{10447454}
Zhang, J., Vahidian, S., Kuo, M., Li, C., Zhang, R., Yu, T., Wang, G., Chen, Y.: Towards building the federatedgpt: Federated instruction tuning. In: ICASSP 2024 - 2024 IEEE International Conference on Acoustics, Speech and Signal Processing. pp. 6915--6919 (2024)

\bibitem{emnlp2024_privateFL}
Zheng, J.Y., Zhang, H.N., Wang, L.X., Qiu, W.J., Zheng, H.W., Zheng, Z.M.: Safely learning with private data: A federated learning framework for large language model. In: Proceedings of the 2024 Conference on Empirical Methods in Natural Language Processing. pp. 5293--5306 (2024)

\end{thebibliography}

\end{document}